\pdfoutput=1

\documentclass[11pt]{article}

\usepackage[final]{neurips}
\usepackage{wrapfig, lipsum, booktabs}
\usepackage{listings}
\usepackage{times}
\usepackage{latexsym}
\usepackage[encapsulated]{CJK}
\usepackage[T1]{fontenc}

\usepackage[utf8]{inputenc}

\usepackage{hyperref}
\usepackage{url}
\usepackage{booktabs}
\usepackage{graphicx}
\usepackage{pgfplots}
\usepackage{amsfonts}
\usepackage{bbold}
\usepackage{algorithmic}
\usepackage{algorithm}
\usepackage{setspace}
\usepackage{multirow}
\usepackage{multicol}
\usepackage{tikz}
\usepackage{subfigure}
\usepackage{arydshln}
\usepackage{textcase}
\title{Mengzi: Towards Lightweight yet Ingenious \\ Pre-trained Models for Chinese}

\author{Zhuosheng Zhang$^{1}$\thanks{Technical report. This work was conducted during the authors' internship at \href{https://langboat.com/}{langboat.com}.} , Hanqing Zhang$^{2*}$, Keming Chen$^{3*}$, Yuhang Guo$^{4*}$, \\ \textbf{Jingyun Hua$^{5}$}, \textbf{Yulong Wang}$^{5}$, \textbf{Ming Zhou}$^{5}$\\
	$^{1}$Shanghai Jiao Tong University, Shanghai, China $^{2}$Beijing Institute of Technology, Beijing, China \\
	$^{3}$Beijing Jiaotong University, Beijing, China $^{4}$Peking University, Beijing, China \\
	$^{5}$Langboat Technology, Beijing, China \\
	\texttt{zhangzs@sjtu.edu.cn,zhanghanqing@bit.edu.cn}\\ 
	\texttt{20120341@bjtu.edu.cn,yuhangguo@pku.edu.cn} \\
	\texttt{\{huajingyun,wangyulong,zhouming\}@chuangxin.com}
}

\begin{document}
\maketitle
\begin{abstract}
Although pre-trained models (PLMs) have achieved remarkable improvements in a wide range of NLP tasks, they are expensive in terms of time and resources. This calls for the study of training more efficient models with less computation but still ensures impressive performance. Instead of pursuing a larger scale, we are committed to developing lightweight yet more powerful models trained with equal or less computation and friendly to rapid deployment. This technical report releases our pre-trained model called Mengzi, which stands for a family of discriminative, generative, domain-specific, and multimodal pre-trained model variants, capable of a wide range of language and vision tasks. Compared with public Chinese PLMs, Mengzi is simple but more powerful. Our lightweight model has achieved new state-of-the-art results on the widely-used CLUE benchmark with our optimized pre-training and fine-tuning techniques. Without modifying the model architecture, our model can be easily employed as an alternative to existing PLMs. Our sources are available at \url{https://github.com/Langboat/Mengzi}.

\end{abstract}

\section{Introduction}
\begin{CJK*}{UTF8}{gkai}
\begin{center}
	\begin{tabular}{p{13cm}}
		\noindent\emph{Using force to suppress others leads to superficial compromise. Genuine power only comes from practicality. (以力服人者, 非心服也,力不赡也。权,然后知轻重;度,然后知长短。)} \\
		\midrule
		\hfill{\emph{Mencius {\rm(372 BC - 289 BC)}}}\\
	\end{tabular}
\end{center}
\end{CJK*}

Pre-trained models (PLMs) have greatly improved performance in a broad spectrum of natural language processing (NLP) tasks and stimulated the development to more practical scenarios \citep{radford2018improving,peters2018deep,devlin-etal-2019-bert,yang2019xlnet,liu2019roberta,lan2019albert,clark2019electra}. 
Various trends have emerged recently: 1) bigger model and more data; 2) more efficient architecture and pre-training methodology; 3) domain- and task-aware pre-training 4) unification of vision and language modeling. With the promising advances above, a variety of pre-trained models have been developed for real-world applications. Despite their convenience of use, PLMs currently consume and require expensive resources and time, which hinders the wide range of practical applications. Therefore, modest-sized but powerful models, i.e., with only 100 million parameters, are much more preferred in light of resource cost and development circle, which desperately calls for the study of efficient methods. From the technical view, the major problems concerning lightweight language models lie within two aspects: effective training objectives that capture knowledge fast and efficient strategies that train language models quickly.

For model effectiveness, although PLMs have shown effectiveness in capturing syntax and semantic knowledge after pre-training \citep{hewitt2019structural,ettinger2020bert}, recent studies show that the current models still suffer from under-fitting issues, and it remains challenging to train a powerful model with less computation \citep{rogers2020primer}. Designing effective criteria for language modeling is one of the major topics in training pre-trained models, which decides how the model captures the knowledge from large-scale unlabeled data. Recent studies have investigated denoising strategies \citep{raffel2020exploring,lewis2020bart}, model architecture \citep{yang2019xlnet}, and auxiliary objectives \citep{lan2019albert,joshi-etal-2020-spanbert} to enhance the model capacity in pre-training. However, the cutting-edge researches mainly focus on English; there are a few studies in other languages like Chinese \citep{wei2019nezha,cui-etal-2020-revisiting,zhang2021cpm,zeng2021pangu}. 
Besides, the application requirements in specific domains, e.g., financial analysis and multimodal tasks, further urge the development of effective Chinese pre-trained models.

To the end of efficiency, recent studies have investigated knowledge distillation \citep{sanh2019distilbert,jiao2020tinybert,wang2020minilm} and model compression techniques \citep{gordon2020compressing,shen2020q,xu2020bert}. However, they are not optimal for real-world applications. Knowledge distillation methods train a light model with the guidance of a large-scale teacher model, which requires two stages of training, and training a teacher model still consumes massive computing resources. Similarly, model compression aims to train a simplified and optimized model from the original one without significantly diminished accuracy. The widely-used techniques include parameter sharing \citep{lan2019albert}, module replacement \citep{xu2020bert}, pruning \citep{gordon2020compressing}, and quantization \citep{shen2020q}. Such a line of methods still needs abundant training. Also, these methods suffer from dramatic changes in the model architecture, so that it would be hard for easy real-world practice as it is incompatible with commonly deployed frameworks like the Transformers toolkit \citep{wolf-etal-2020-transformers}.
\begin{wrapfigure}{r}{0.5\textwidth}
  \begin{center}
    \includegraphics[width=0.5\textwidth]{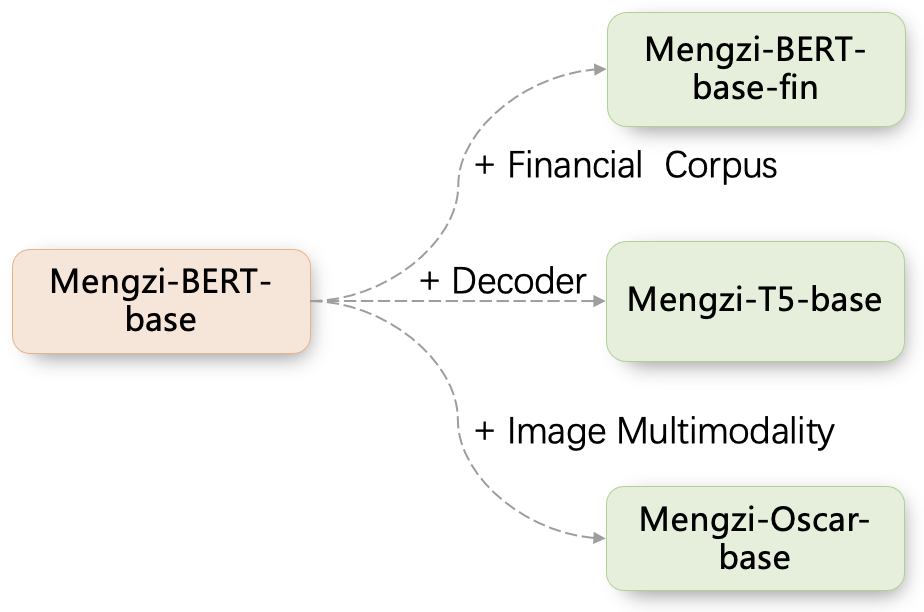}
  \end{center}
  \setlength{\abovecaptionskip}{0pt}
  \caption{The family of Mengzi models. Mengzi-BERT-base-fin, Mengzi-T5-base, and Mengzi-Oscar-base are derivatives of Mengzi-BERT-base.}
	\label{fig:derivative}
\end{wrapfigure}


In this work, instead of pursuing larger model size as the major goal of recent studies, we aim to provide more powerful but much resource-friendly models with a better performance compared with others on the same scale, which are of potential to rapid application to real scenarios and large-scale deployment. Therefore, we seek carefully optimized enhancement on the pre-training objectives, inspired by linguistic analysis and training acceleration, and are also free from a model architecture modification. As a result, we develop Mengzi, which is a family of discriminative, generative, domain-specific, and multimodal pre-trained model variants capable of a wide range of language and vision tasks. To keep consistent with public models and ensure easy application, we build our backbone model on top of the RoBERTa \citep{liu2019roberta} following the same model settings. The main contributions of this work are three-fold:

1) We investigate various pre-training strategies to train lightweight language models, showing that well-designed objectives can further significantly improve the model capacity without the need to enlarge the model size.

2) We release Mengzi, including the discriminative, generative, financial, and multimodal model variants, capable of a wide range of language and vision tasks. The text encoders in these models only contain 103 million parameters, which we hope to facilitate the related studies for both academia and industry.
 
3) Extensive evaluates on widely-used benchmarks demonstrate that Mengzi achieves strong performance on a range of language understanding and generation tasks.

\section{Backbone Encoder}\label{sec:encoder}
Figure \ref{fig:derivative} shows the family of released Mengzi models and their connections: Mengzi-BERT-base, Mengzi-BERT-base-fin, Mengzi-T5-base, and Mengzi-Oscar-base. From the perspective of the application scenario, they range from text-only language models to multimodal variants, from general-purpose training to domain-specific adaptation.
The details will be demonstrated in Section \ref{sec:release}. From a technical point of view, the last three ones can be regarded as the derivatives of Mengzi-BERT-base because their text encoders follow the same structure as Mengzi-BERT-base and are initialized by the pre-trained parameters of Mengzi-BERT-base. Therefore, in the following experimental parts, for simplicity, we only focus on the fundamental text-only encoder side and report our optimization techniques that are of general effectiveness.\footnote{We denote Mengzi-BERT-base as Mengzi for short in subsequent parts unless otherwise specified.}

\subsection{Setup}
Data, algorithms, and computation are the key to powerful pre-trained language models. In the following part, we will present the details for training Mengzi in view of the three aspects.
\paragraph{Data Processing}
The pre-training corpus is derived from Chinese Wikipedia, Chinese News, and Common Crawl, with a 300GB data size in total. We clean the data by using exploratory data analysis techniques, i.e., removing HTML tags, URLs, e-mails, emoji, etc. Since there are simplified and traditional Chinese tokens in the original corpus, we convert traditional tokens into the simplified form using OpenCC.\footnote{\url{https://github.com/BYVoid/OpenCC}.} Duplicate articles are also removed.

\paragraph{Architecture}
RoBERTa \citep{liu2019roberta} is leveraged as the initial backbone model for Mengzi pre-training. Our Mengzi architecture is based on the \texttt{base} size, where the model consists of 12 transformer layers, with the hidden size of 768, 12 attention heads, and 103M model parameters in total. We keep the model specification the same as the public one to ensure compatibility in real-world deployment and application. Following \citet{liu2019roberta}, we employ masked language modeling (MLM) as the major pre-training task.
\paragraph{Pre-training Details}
Our vocabulary contains 21,128 tokens. We limit the length of sentences in each batch to up to 512 tokens, and the batch size is 128. During pre-training, 15\% words are randomly masked in each sequence for MLM prediction.  We use a mixed-batch training procedure with LAMB optimizer \citep{you2019large}, which involves two stages: the first 9/10 of the total epochs use a sequence length of 128, and the last 1/10 of the total epochs use a sequence length of 512. The batch sizes for the two stages are 16384 and 32768, respectively. We employ PostgreSQL to globally sample the training examples to avoid the imbalance of sample weight in the two-stage training. The overall pre-training process takes 1,000,000 steps. We use 32 NVIDIA Tesla 24GB 3090 Ti GPUs, with FP16 and deepspeed\footnote{\url{https://github.com/microsoft/DeepSpeed}.} for training acceleration.

\section{Experiments}
\subsection{Tasks}
For downstream tasks for model evaluation, we use the Chinese Language Understanding Evaluation (CLUE) benchmark \citep{xu2020clue}, which consists of six different natural language understanding tasks: Ant Financial Question Matching (AFQMC), TouTiao Text Classification for News Titles (TNEWS), IFLYTEK \citep{co2019iflytek}, Chinese-translated Multi-Genre Natural Language Inference (CMNLI),  Chinese Winograd Schema Challenge (WSC), and Chinese Scientific Literature (CSL) and three machine reading comprehension (MRC) tasks: Chinese Machine Reading Comprehension (CMRC) 2018 \citep{cui-etal-2019-span},  Chinese IDiom cloze test (CHID) \citep{zheng-etal-2019-chid}, and Chinese multiple-Choice machine reading Comprehension (C$^3$) \citep{sun2019probing}.

\subsection{Setup}
We build the downstream models for the natural language understanding tasks by adding a linear classifier on top of the ``{\tt [CLS]}" token to predict label probabilities. For the span-based question answering task, CMRC, we packed the question and passage tokens together with special tokens to form the input: ``{\tt [CLS]} Question {\tt [SEP]} Passage {\tt [SEP]}'', and employed two linear output layers to predict the probability of each token being the start and end positions of the answer span following the practice for BERT \citep{devlin-etal-2019-bert}. 
For multi-choice MRC tasks, CHID and C$^3$, we concatenated the passage, question, and each candidate answer (``{\tt [CLS]} Question $||$ Answer {\tt [SEP]} Passage {\tt [SEP]}''), then predicted the probability of each answer on the representations from the ``{\tt [CLS]}'' token following prior works \citep{yang2019xlnet, liu2019roberta}.

\subsection{Implementation Details}
For the fine-tuning experiments, we use Adam as our optimizer with an initial learning rate in \{8e-6, 1e-5, 2e-5, 3e-5\} with a warm-up rate of 0.1 and L2 weight decay of 0.01. The batch size is selected in \{16, 24, 32\}. The maximum number of epochs is set in [2, 5] depending on tasks. Texts are tokenized with a maximum length of 384 for MRC and 256 for other tasks. 

\begin{table*}
	\centering
	\setlength{\tabcolsep}{2.1pt}
	\begin{tabular}{ l c c c c cc c c cc c c}
	    \toprule
		{\bf Models} & {\bf Scale} &\multicolumn{3}{c}{{\bf Sentence-Pair}} & & \multicolumn{3}{c}{{\bf Single-Sentence}} & & \multicolumn{3}{c}{{\bf MRC}} \\
		\cmidrule{3-5} \cmidrule{7-9} \cmidrule{11-13} &  & AFQMC& CMNLI  & CSL & & TNEWS  &  IFLYTEK&  WSC  & &  CMRC18 &  CHID & C$^3$	\\
		\midrule
		\multicolumn{12}{l}{\textit{Single-task single models on dev (\texttt{base} models)}}\\
		BERT &  108M & 74.16 & 79.47 & 79.63  & & 56.09 & 60.37 & 59.60 & & 75.13 & 82.20 & 65.70 	\\
		RoBERTa & 108M &74.30 & 80.70 & 80.67 & & 57.51 & 60.80 & 67.20 & & 77.59 & 83.78 & 67.06 \\
		Mengzi & 103M &74.58 & 82.12 & 85.40 & & 57.97 & 60.68 & 87.50 & & 78.54 & 84.16 & 71.70 \\
		\midrule
		\multicolumn{12}{l}{\textit{Official leaderboard results on test (\texttt{large} models with enhancements)}}\\
		Pangu & 200B & 78.11 & 85.19 & 87.73  && 72.07 & 65.19 & 95.52 & & 84.45 & 93.25 & 85.64 \\
		BERTSG & \textasciitilde10B &79.85 & 85.30 & 89.00  && 74.15 & 64.54 & 95.17  && 83.80 & 93.06 & 87.44 \\
		Motian & \textasciitilde1B &78.30 & 85.44 & 90.17 &  & 73.18 & 65.46 & 94.83 &  & 85.30 & 94.42 & 88.49 \\
		ShenZhou & \textasciitilde10B &80.29 & 86.49 & 90.97 & & 74.15 & 67.65 & 95.17 & & 85.30 & 94.42 & 88.49 \\
		Mengzi & \textasciitilde1B &\textbf{81.79} & \textbf{86.13} &  \textbf{89.87} & & \textbf{75.06} & \textbf{65.08} & \textbf{96.55} &&  \textbf{83.95} &  \textbf{96.00} & \textbf{92.39} \\	
		\bottomrule
	\end{tabular}   
	\caption{Results on the CLUE development datasets. The RoBERTa Dev results is from \citet{cui-etal-2020-revisiting}. The test results except ours are from the CLUE leaderboard. Since there is a lack of accurate numbers of parameters in some public models, we use \textasciitilde to indicate the approximate scale. The standard evaluation metric is accuracy. For CMRC18, the reported score is calculated by the average of EM and F1 scores.}
	\label{table:clue}
\end{table*}

\subsection{Overall Results}
Table \ref{table:clue} shows the performance of Mengzi on CLUE compared with pubic models. Compared with the RoBERTa baseline, we observe that Mengzi achieves consistent improvements on all the subtasks, showing that Mengzi is an effective alternative. For the public ranking on the test set, our large model has surpassed existing models for over three months. Mengzi not only  far exceeds the performance of  public models under the same model scale but also outperforms the largest Chinese model with 200 billion parameters, Pangu \citep{zeng2021pangu}.\footnote{The large model follows the \texttt{large} setting in \citet{liu2019roberta} and uses the same pre-training process as our \texttt{base} model.}

Taking the Mengzi model as the backbone, we are interested in whether extra plug-in techniques, like auxiliary training objectives, would further improve the model capacity. In view of industrial deployment, assume that once the PLM is deployed, we would not spare manual labor to update the environment or model framework. The simplest way is to update the existing model weights with a new one. Therefore, we keep the basic criteria that those techniques should be independent of the model architecture, beneficial for pre-training, and dispensable during inference. To this end, we investigate the pre-training and fine-tuning techniques to enhance the capacity of Mengzi further.

\section{Analysis}
\subsection{Pre-training Techniques}
\paragraph{Linguistic-motivated Objectives} 
Linguistic information has been shown effective for language modeling \citep{xu-etal-2021-syntax,zhang2020sg}. Inspired by LIMIT-BERT \citep{zhou2020limit}, we employ part-of-speech (POS) and named entity (NE) sequence labeling tasks in conjunction with the original MLM and NSP objective during pre-training. POS and NE tags in the raw texts are annotated by spaCy.\footnote{\url{https://github.com/explosion/spaCy}.} 

\paragraph{Sequence Relationship Objectives}
To better model the pairwise information between sentences, we add the sentence order prediction (SOP) task \citep{lan2019albert} in model pre-training. 

\paragraph{Dynamic Gradient Correction}
The widely-used MLM would cause the disturbance of original sentence structure, leading to the loss of semantics and improve the difficulty of prediction, inevitably resulting in insufficient and inefficient training. To alleviate the issue, we propose a series of dynamic gradient correction techniques to improve the model capacity, as well as the robustness.\footnote{More details will be provided in our latter version.} 


\begin{wraptable}{r}{0.5\textwidth}
\setlength{\tabcolsep}{8pt}
\begin{tabular}{lc}
    \toprule
    {\bf Models}    & {\bf Accuracy} (\%)\\
    \midrule
    Baseline & 81.4 \\
    \quad + Knowledge Distillation  & 82.6\\
    \quad + Data Augmentation  & 85.3\\
    \bottomrule
  \end{tabular}
  \caption{Ablation results on the CMRC2018 dev set (average accuracy of F1 and EM scores).}\label{ablation-cmrc}
\end{wraptable}

\subsection{Fine-tuning Strategies}
Fine-tuning strategies are essential for downstream task performance. We report the results of the general and representative techniques that we have investigated, including knowledge distillation, transfer learning, choice smoothing, adversarial training, and data augmentation to further enhance the fine-tuning performance. Since those strategies mainly aim for competing on the leaderboard, the analysis is based on \texttt{large} models.

\begin{wraptable}{r}{0.5\textwidth}
\setlength{\tabcolsep}{8pt}
\begin{tabular}{lc}
    \toprule
    {\bf Models}    & {\bf Accuracy} (\%)\\
    \midrule
    Baseline & 75.2  \\
    \quad + Knowledge Distillation  & 77.1 \\
    \quad + Transfer Learning & 77.3 \\
    \quad + Choice Smoothing  & 77.8 \\
    \bottomrule
  \end{tabular}
  \caption{Ablation results on the C$^3$ dev set.}\label{ablation-c3}
\end{wraptable}

\paragraph{Knowledge Distillation} We train a teacher model and employ the teacher model to guide the training of the student model. In detail, we calculate the Kullback–Leibler (KL) divergence of the contextualized hidden states from the teacher and student models, respectively, for the same input sequence. The divergence measures the similarity degree between the representations from the teacher and student models, which is minimized during fine-tuning, along with the original downstream task objective.

\paragraph{Transfer Learning} We leverage the parameters from the trained model on the CMNLI dataset to initialize the model training for related datasets like C$^3$. For AFQMC, we use the model trained on LCQMC \citep{liu2018lcqmc} and XNLI (the Chinese part) \citep{conneau2018xnli} and initialize the model training on AFQMC. For CMNLI, we first use the OCNLI \citep{hu2020ocnli}, CMNLI, SNLI \citep{bowman2015large}, MNLI (translated) \citep{nangia2017repeval}, and XNLI (Chinese part) \citep{conneau2018xnli} for training an initial model, and then use it for initializing CMNLI model training.

\paragraph{Choice Smoothing} For multi-choice or classification tasks, combining different kinds of training objectives would lead to better performance \citep{zhang2021retro}. For each input example, we apply the cross-entropy and the binary cross-entropy as the loss functions and combine the loss from both sides to help the model learn features from different granularity.

\paragraph{Adversarial Training} To help the model generalize to unseen data, we apply a smoothness-inducing adversarial regularization technique following \citet{jiang2020smart} to encourage the output of the model not to change much when injecting a small perturbation to the input.

\begin{wraptable}{r}{0.5\textwidth}
\setlength{\tabcolsep}{8pt}
\begin{tabular}{lc}
    \toprule
    {\bf Models}    & {\bf Accuracy} (\%)\\
    \midrule
    Baseline & 85.6  \\
    \quad + Choice Smoothing  & 85.8 \\
    \quad + Adversarial Training & 86.7 \\
    \quad + Data Augmentation  & 88.4 \\
    \bottomrule
  \end{tabular}
  \caption{Ablation results on the CHID dev set.}\label{ablation-chid}
\end{wraptable}
\paragraph{Data Augmentation} Data augmentation has been widely used for training powerful models, especially for low-resource situations. For tasks like CHID, for each idiom in the given dictionary, we collect the related sentences from our corpora for pre-training and use them as extra training sources. For CMRC2018, we add the training data from DRCD and SQuAD (translated) for augmentation. In addition, we use the original version in CLUEWSC2020 as supplemental training set for training WSC models.\footnote{\url{https://github.com/dbiir/UER-py/wiki/CLUE-Classification}.} 



Tables \ref{ablation-cmrc}-\ref{ablation-chid} show the ablation results of the representative fine-tuning strategies, from which we have the following observations:

1) For MRC tasks like CMRC2018 and C$^3$, knowledge distillation can boost the benchmark performance of the student model with the guidance of teacher predictions.

2) Transfer learning boosts the model performance, which is consistent with practice for English GLUE benchmark \citep{wang-etal-2018-glue} by using Multi-Genre Natural Language Inference (MNLI) \citep{N18-1101} for initialization \citep{liu2019roberta,lan2019albert} that would be beneficial for training models on small-scale datasets. However, we find that such a way of transfer learning is also helpful for large datasets like C$^3$.

3) Choice smoothing is effective for multi-choice tasks, which may provide fine-grained information from multi-label classification and binary classification, where multi-label classification captures the relationship between the label-wise predictions and binary classification models the prediction confidence for each label.

4) Adversarial training shows obvious improvements on CHID, which might be due to the benefit that using small perturbations in the embeddings might help improve the model's robustness.

5) Data augmentation is an effective approach to enhance the model capacity. We observe substantial improvements in CMRC and CHID. However, finding a suitable augmentation technique remains a challenge.

\begin{table}
 \setlength{\tabcolsep}{6pt}
  \setlength{\tabcolsep}{3.8pt}

\begin{tabular}{p{0.12\textwidth}p{0.05\textwidth}p{0.28\textwidth}p{0.26\textwidth}p{0.2\textwidth}}%
\toprule
\textbf{Model} & \textbf{Size} & \textbf{Features} & \textbf{Tasks} & \textbf{Corpus} \\
\midrule
Mengzi-BERT-base & 103M & Compatible with BERT as a stronger alternative, powered with linguistic-driven enhancements.  & Text classification, entity recognition, relation extraction, reading comprehension, etc. & 300G internet corpus \\
\midrule
Mengzi-T5-base & 220M &  More controllable text generation capacity, better performance than BERT structure and GPT structure. & Article generation, news generation, financial research report generation, etc. &  300G internet corpus  \\
\midrule
Mengzi-BERT-base-fin & 103M &  Specific for financial tasks by training Mengzi-BERT-base with financial corpus.  &  Financial news classification, sentiment analysis of financial research reports, etc. & 20G financial news, announcements, research reports \\
\midrule
Mengzi-Oscar-base & 103M & Applicable to multimodal tasks, on top of Mengzi-BERT-base trained on millions of text-image pairs  & image caption, text retrieval, image retrieval, etc. & 300G internet corpus + 3.7M Chinese text-image pairs  \\
\bottomrule
\end{tabular}
\caption{Details of the release Mengzi models.}
\label{tab:mis_example}
\end{table}

\begin{figure*}[htb]
	\centering
	\includegraphics[width=1.0\textwidth]{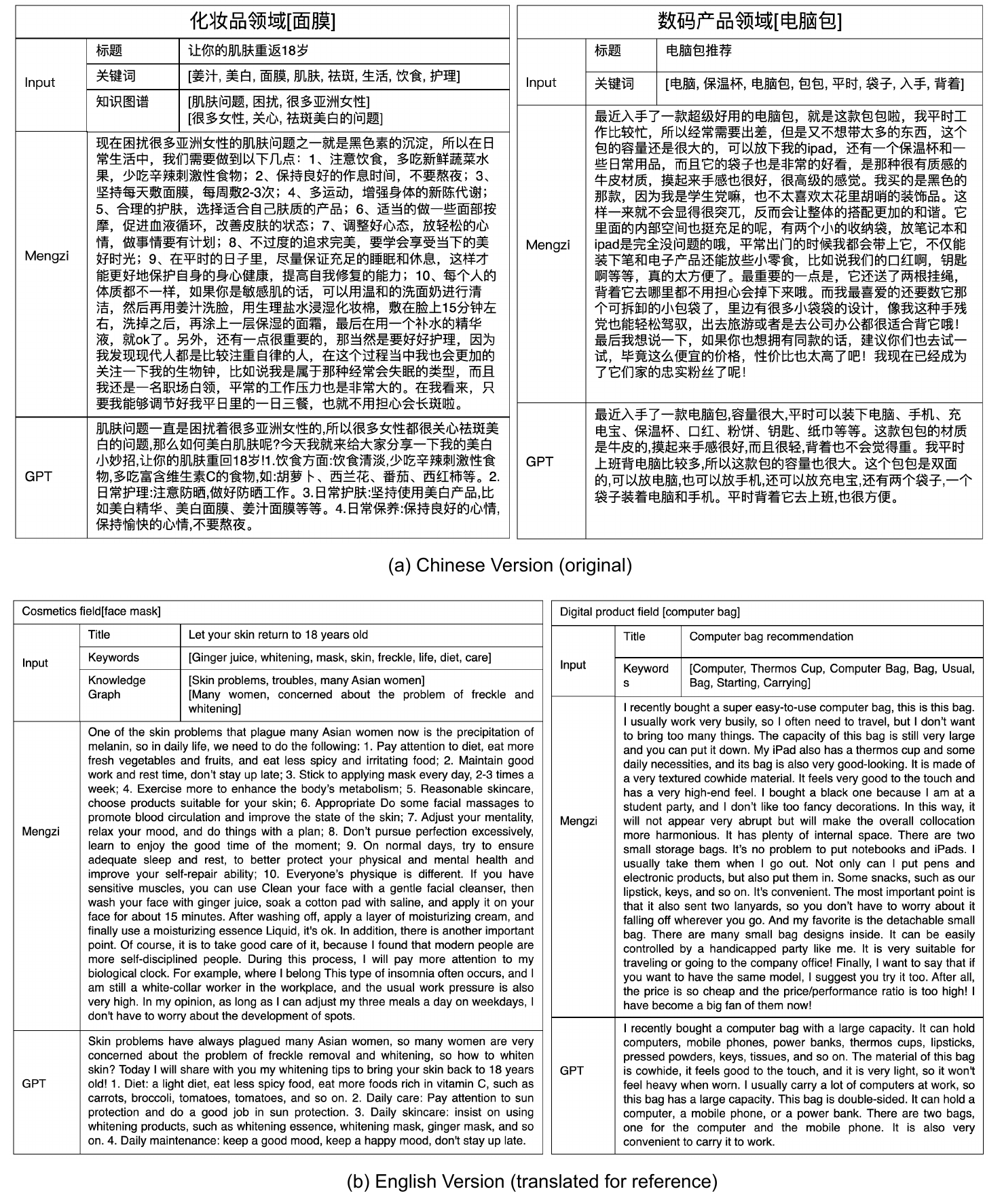}
	\caption{Generated marketing copywriting examples from Mengzi-T5-base and GPT.}
	\label{fig:fin}
\end{figure*}

\section{Model Release}\label{sec:release}
We release a family of pre-trained models covering discriminative, generative, multimodal, and financial application areas on the backbone of our ingenious encoders. The details of the release Mengzi models are presented in Table \ref{tab:mis_example}. Mengzi-BERT-base initializes the text encoders of Mengzi-BERT-base-fin, Mengzi-T5-base, and Mengzi-Oscar-base.

\paragraph{Mengzi-BERT-base} is a discriminative language model compatible with BERT as described in Section \ref{sec:encoder}, which can be used for most NLP tasks like natural language understanding and machine reading comprehension.

\paragraph{Mengzi-T5-base} is a generative language model with a decoder module specialized for natural language generation tasks. The overall architecture follows T5 \citep{raffel2020exploring}.

\paragraph{Mengzi-BERT-base-fin} is a domain-specific language model designed for financial scenarios, by continuing training Mengzi-BERT-base using our collected 20G financial corpus composed of financial news, announcements, and financial research reports.

\paragraph{Mengzi-Oscar-base} is a multimodal model effective for vision-language tasks, like image caption. The overall architecture follows Oscar \citep{li2020oscar,zhang2021vinvl}, which is a  vision-language pre-training method to learn generic image-text representations for vision-language understanding and generation tasks. The language encoder is initialized by our Mengzi-BERT-base.

\begin{figure*}
	\centering
	\includegraphics[width=1.0\textwidth]{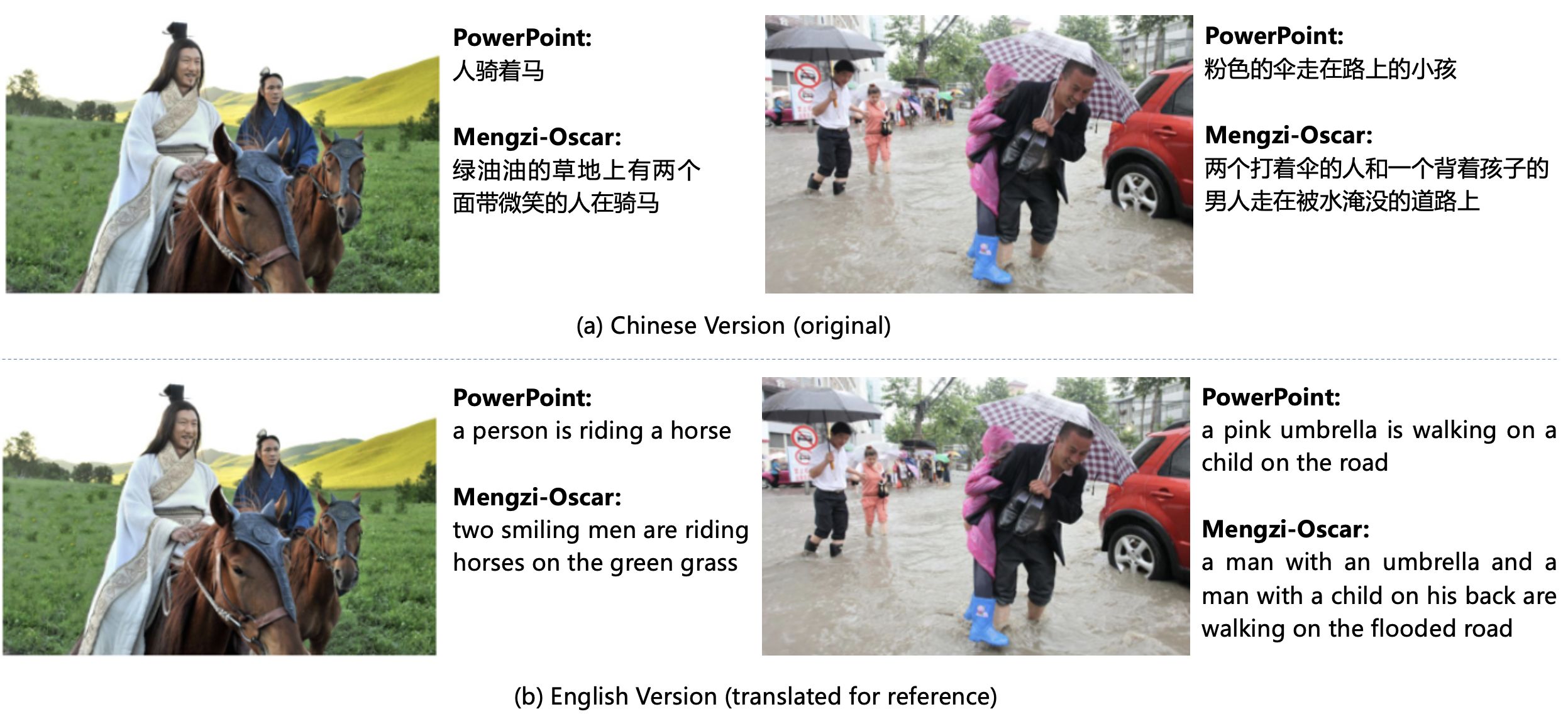}
	\caption{Generated caption examples from Mengzi-Oscar-base and PowerPoint (Randomly selected from the AIC-ICC val set).}
	\label{fig:multi}
\end{figure*}

\begin{table*}[t]
	\centering
	\setlength{\tabcolsep}{2.2pt}\small
	{
\begin{tabular}{lcccc}
    \toprule
    {\bf Models}    & {\bf Information Retrieval} & {\bf Entity Recognition} & {\bf Relation Extraction} & {\bf Entity Linking} \\
    \midrule
    RoBERTa-wwm-ext & 90.20/92.90 & 88.11 & 77.44 & 93.40  \\
    Mengzi-BERT-base  & 90.40/92.40 & 88.51 & 77.51 & 93.80 \\
    Mengzi-BERT-base-fin & 91.00/93.50 & 88.53 & 77.57 & 94.10 \\
    \bottomrule
  \end{tabular}
    }
	\caption{Experimental results in the financial domain. The RoBERTa-wwm-ext baseline is from \citet{cui-etal-2020-revisiting}. The evaluation results of information retrieval are reported by R@10/20. For entity recognition and relation extraction, the metric is F1. For entity linking, we use accuracy.}\label{ana-fin}
	\label{maintable}
\end{table*}

\subsection{Exemplars and Assessment}

\paragraph{Marketing Copywriting} 
Figure \ref{fig:fin} compares the quality of the generated marketing copywriting texts based on our Mengzi-T5-base model and GPT. Given the input title and keywords, the models are required to generate a corresponding descriptive passage. According to the generated examples, we observe that texts generated by our  Mengzi-T5-base model contain much more details and keep fluency at the same time, indicating that generating texts using our model would benefit from satisfactory diversity fluency and coherence.

\paragraph{Financial Tasks}
We evaluate our Mengzi-BERT-base and Mengzi-BERT-base-fin in financial tasks, such as information retrieval, entity recognition, relation extraction, and entity linking. We extract the entities (e.g., events) from LUGE for the entity recognition task.\footnote{https://aistudio.baidu.com/aistudio/competition/detail/46/0/task-definition.} For evaluation on the other tasks, we use our self-collected datasets. Results in Table \ref{ana-fin} show that our methods are capable of the tasks specific for the financial domain, especially our Mengzi-BERT-base-fin yields the best performance.

\paragraph{Image Caption}
We compare the image caption performance of Mengzi-Oscar-base with the widely-used Automatic Alt Text technique used in Microsoft 365.\footnote{https://support.microsoft.com/en-us/topic/everything-you-need-to-know-to-write-effective-alt-text-df98f884-ca3d-456c-807b-1a1fa82f5dc2.} Figure \ref{fig:multi} shows the case studies based on randomly selected examples from the AIC-ICC Val set \citep{wu2017ai}. We observe that our model generates more fluent and informative captions compared with the baseline.

\subsection{How To Use}
Our released Mengzi models are available at \url{https://github.com/Langboat/Mengzi}. 
Our models are also easily accessible by using the HuggingFace Transformers toolkit.\footnote{\url{1https://huggingface.co/mengzi}.} For example, Mengzi-BERT-base is available by calling through the following scripts:

\begin{lstlisting}
from transformers import BertTokenizer, BertModel
tokenizer = BertTokenizer.from_pretrained("Langboat/mengzi-bert-base")
model = BertModel.from_pretrained("Langboat/mengzi-bert-base")
\end{lstlisting}

\section{Conclusion}
This technical report presents our exploration of training lightweight language model called Mengzi, which shows remarkable performance improvements compared with the same-sized or even larger-scale models. A series of pre-training and fine-tuning strategies have been verified to be effective for improving model benchmark results. Experimental results show that Mengzi achieves state-of-the-art performance with carefully designed training strategies. Without the modification of the model architecture, Mengzi is easy to be deployed as a powerful alternative to existing PLMs. 

\bibliography{custom}
\bibliographystyle{acl_natbib}

\appendix

\end{document}